\documentclass[a4]{article}
\usepackage{graphicx}
\usepackage{algorithm2e}

\def\be{\begin{equation}}
\def\ee{\end{equation}}
\newcommand{\ff}[1]{{\bf  #1}}
\def\a{\alpha}

\def\lam{\lambda}
\def\x{\ff{x}}

\title{Cuckoo Search: State-of-the-Art and Opportunities}

\author{Xin-She Yang \\
School of Science and Technology,\\
Department of Design Engineering and Mathematics, \\
Middlesex University,  London NW4 4BT, UK.
\and
Suash Deb \\
 IT \& Educational Consultant,
Ranchi, India \\
 Distinguished Professorial Associate, \\
Decision Sciences and Modelling Program, \\
Victoria University, Melbourne, Australia.   }

\begin{document}

\maketitle

\begin{abstract}
Since the development of cuckoo search (CS) by Yang and Deb in 2009, CS has been applied in a diverse range of applications. This paper first outlines the key features of the algorithm and its variants, and then briefly summarizes the state-of-the-art developments in many applications. The opportunities for further research are also identified.
\end{abstract}

\noindent {\bf Citation detail:}
X.-S. Yang, S. Deb, Cuckoo search: State-of-the-art and opportunities, Proceedings of the 4th IEEE Conference on Soft Computing and Machine Learning (ISCMI2017), 23-24 Nov 2017, Port Louis, Mauritius, pp. 55-59 (2017). https://doi.org/10.1109/ISCMI.2017.8279597


\section{Introduction}
Nowadays metaheuristic algorithms are widely used for solving optimization
problems related to engineering designs, data mining, machine learning
and image processing. These algorithms are usually simple and flexible, ease for implementation and yet effective in practice \cite{YangDebRev}. Despite their success, there
are still some areas that are less explored and applied and thus further research opportunities exist in many areas.
As there are so many metaheuristic algorithms, it is not possible to address
all the challenges in general. Instead, we focus on the applications and
opportunities concerning cuckoo search (CS) and its variants.

Therefore, the paper is organized as follows. Section II provides an outline of the standard cuckoo search and its variants.  Section III highlights some most recent applications,
and Section IV points out a few further research opportunities.

\section{Cuckoo Search and Applications}

\subsection{Cuckoo Search}

Cuckoo search (CS), developed in 2009 by Xin-She Yang and Suash Deb \cite{YangDeb},
is a bio-inspired metaheuristic algorithm. It uses the key features of
the brood parasitism of some cuckoo species and their co-evolution with
host bird species \cite{Davi}. In the standard cuckoo search, one crucial simplification
is that a cuckoo lays only one egg which represents a solution vector
and each nest can only have one egg. In this case,  there is no distinction
between an egg, a nest, or a cuckoo \cite{YangDeb2010}.

For simplicity, we use $\x=(x_1, x_2, ..., x_D)$ to represent a $D$-dimensional vector
for an optimization problem with an objective function $f(\x)$ subject to various constraints. The main equation in CS to simulate the egg-laying behaviour and similarity
of solutions (eggs) can be written as
\be \x_i^{t+1}=\x_i^t +\beta s \otimes H(p_a-\epsilon) \otimes (\x_j^t-\x_k^t), \label{equ-100}  \ee
where $\x_j^t$ and $\x_k^t$ are two different solutions selected randomly by random permutation. Here, the discovery probability $p_a$ of a cuckoo egg by a host bird is realized by the Heaviside function ($H(u)=1$ if $u>0$ and $H(u)=0$ if $u<0$). In addition,
$\epsilon$ is a uniformly distributed random number in [0,1] and
$s$ is the step size. Here $\beta$ is a scaling factor.

The other equation uses L\'evy flights to mimic the locations of nests to be built by hosts
\be \x_i^{t+1}=\x_i^t+\a \otimes L(s, \lam), \label{equ-200} \ee
with \be L(s, \lam) \sim \frac{\lam \Gamma(\lam) \sin (\pi \lam/2)}{\pi}
\Big(\frac{1}{s^{1+\lam}}\Big), \quad (s \gg 0), \ee
where `$\sim$' means that the random numbers $L(s,\lam)$ should be drawn from the L\'evy distribution that is approximated by a fat-tailed distribution such as
a power-law distribution with an exponent $\lam$.
The parameter $\a>0$ is the step size scaling factor determined by the
scales or bound ranges of the problem of interest. From the implementation  point of view,  $\otimes$ is an entry-wise operator.

In comparison with the randomization techniques used in other algorithms such as particle swarm optimization, CS uses a more sophisticated L\'evy distribution to carry out L\'evy flights. As L\'evy distribution is a fat-tailed distribution, the steps generated can have
both large and small components, which enables the cuckoo search to do both large-scale exploration as well as local exploitation. Such randomization can also make the algorithm more likely to jump out of any local optima, enabling a better overall exploration ability as well as a sufficiently strong exploitation ability \cite{YangDeb,YangDebRev}. CS has also been extended to solve multiobjective optimization \cite{YangDeb2013,Kaveh}.

\subsection{Variants}

Many variants of cuckoo search have been designed by researchers by introducing additional components or by hybridization with other optimization techniques. Shehab et al. provided a comprehensive review \cite{Shehab}
about such variants, and thus we only highlight a few variants here.

\begin{itemize}

\item Binary cuckoo search: A binary cuckoo search was developed to carry out feature selection \cite{Perei}.

\item Chaotic cuckoo search: A chao-enhanced cuckoo search was developed for global optimization \cite{Huang}, which used various chaotic maps to carry out randomization.
    Another variant of chaotic cuckoo search has been developed by Wang et al. \cite{WangG}.

\item Modified cuckoo search: A modified cuckoo search was developed and used to carry out aerodynamic shape optimization \cite{Naum}. A snap-drift cuckoo search was developed with snap and drift learning strategy \cite{Rakh}, while a nearest-neighbour cuckoo search was developed by using neighbour properties \cite{Wang}.

\item Self-adaptive cuckoo search: A self-adaptive cuckoo search was developed with adaptive parameter values \cite{Mlak}, while a modified cuckoo search with self-adaptive parameter was designed using mutation rules and varying parameters \cite{Li}.

\item Multiobjective cuckoo search: A multiobjective cuckoo search was developed to solve design optimization problems with multiples objectives \cite{YangDeb2013}.

\end{itemize}

There are other variants, and readers can refer to \cite{Shehab}. Here, the rest of this paper will focus on the recent applications.

\section{Applications}

The applications of cuckoo search are very diverse and the literature is expanding.
A quick Google search gives several thousand entries, while a keyword search in
ScienceDirect shows that there are more than 1000 research papers in the past few years.
So it is not possible to review even a small fraction of these studies.

Though a relatively comprehensive review of the studies was carried out by Yang and Deb \cite{YangDebRev}, however, the literature was covered up to 2014.  In addition, other reviews of cuckoo search were carried out by Fister et al.~\cite{Fister} and Mohamad et al.~\cite{Moham}, including discussions of various variants. A recent survey was
done by Shehab et al. \cite{Shehab}. An edited book contains many more applications of cuckoo search \cite{YangCSFA}. Therefore, our emphasis here is placed on some representative applications, especially the most recent studies so as to show the diversity of these applications. Obviously, there are many areas of application, we have to select a few area as follows:

\subsection{Design Optimization}

Many applications concern the optimization of a system or its important parts, and researchers have used cuckoo search and its variants to carry out optimization such as optimal positioning platform \cite{Dao}, structural optimization \cite{Gandomi}, affinity propagation \cite{Jia}, aerodynamic shape optimization \cite{Naum}, optimization related to wellbore trajectories \cite{Wood}, dimensionality reduction \cite{Yamany}, overhead crane system optimization \cite{Zhu} and parameter estimation in biological systems \cite{Rakh2}.

In addition, the maximum power point tracking for a PV system \cite{Ahmed} and a PV system with partial shading \cite{Shij} was carried out by cuckoo search. The optimal chiller load was achieved for energy conservation \cite{Coel}. Furthermore, parameter estimation and system identification have been carried out concerning fuzzy systems \cite{Ding}, nonlinear system identification \cite{Gotm},  nonlinear parameter estimation models for ionic liquids \cite{Jaime} and the extraction of T-S fuzzy models \cite{Turki}.

Path planning for mobile robots was done successfully \cite{Mohan}. Multiobjective design optimization was also investigated with various case studies by multiobjective cuckoo search \cite{Kaveh}, and  the fixture layout of metal parts was optimized by cuckoo search \cite{Yangb}.

\subsection{Economic Load Dispatch and Power System}

Another important area of current research is the application related to power systems, including the economic load dispatch \cite{Basu},  the design of power system stabilizers \cite{Abd}, frequency regulation in a wind power system \cite{Chaine} and solar-hybrid cogeneration cycle optimization \cite{Khosh}.

Optimal designs and performance enhancement were also carried out concerning the integrated power system for energy self-sufficient farms \cite{Piech}, the performance of a biodiesel engine \cite{Wong} and PV/wind systems for diary farms \cite{Nadj}. In addition, refueling cycle was improved with thermo-neutronic solvers \cite{Yariz}.

\subsection{Combinatorial Optimization}

Combinatorial optimization tends to be very challenging to solve and in recent years cuckoo search has been applied to solve scheduling problems and resource allocation as well as the travelling salesman problem. Current applications include flowshop scheduling \cite{Dasgu}, control allocation of aircraft \cite{Lu}, multi-skilled multi-objective project scheduling \cite{Magh}, 2-machine robotic cell scheduling \cite{Maju}, random-key cuckoo search for travelling salesman problem \cite{Ouaa} and resource optimization of datacenters \cite{Sait}.

Hybrid flow shop scheduling problems was solved using cuckoo search \cite{Marich},
 and other scheduling problems solved include short-term hydrothermal scheduling \cite{Nguy}, distribution network reconfiguration \cite{Nguy2}, and flow shop scheduling by cuckoo search with hybrid strategies \cite{Wangh}.

\subsection{Filter Design and Traffic Control}

Optimal design of filters and controllers is relevant to many applications such as linear phase multi-band filters \cite{Dash}, higher-order two-channel filter bank \cite{Dhabal}, traffic signal controller \cite{Araghi}, fractional order PID control design \cite{Zamani},
fractional delay-IIR filter design \cite{Kumar} and recursive filter design \cite{Saran}.

\subsection{Time Series and Forecasting}
Forecasting and time series were traditionally done using traditional time series techniques with some limitations. Nowadays, researchers tend to use metaheuristic algorithms in combination with other techniques to do better forecasting. Cuckoo search in particular has been applied to do fractal interpolation for estimating missing values in time series \cite{Jiang},  electrical power forecasting \cite{Xiao}, multi-factor high-order stock forecasting \cite{Zhang}, short-term electrical load forecasting \cite{Zhang2}.

In addition, modelling and predications were also carried out about daily PM$_{2.5}$ concentration prediction \cite{Sunw}, forecast of the number of tourists \cite{Sunx} and solar radiation \cite{Wangj}.

\subsection{Data Mining}

Another increasingly popular and important area is data mining. Cuckoo search has been used to carry out clustering of web search results \cite{Cobos}, medical data classification \cite{Moha}, twitter sentiment analysis \cite{Pandey}, email spam classification \cite{Kumare}, multi-document summarization \cite{Raut} and information granule formation \cite{Sanch}.
In addition, CS has also been used to do feature extraction for graphic objects \cite{Wozniak}. A set of applications in life science data mining were carried out by Fong et al. \cite{Fong} with various swarm techniques.

\subsection{Signal and Image Processing}

Recent applications of cuckoo search have been successful in enhancing performance in image processing techniques. For example, CS has been applied to face recognition \cite{Naik} and image segmentation such as color image segmentation \cite{Nandy}, multilevel color image segmentation \cite{Pare}, and multilevel thresholding for satellite image segmentation \cite{Suresh}. CS has also been used to design the optimum wavelet mask for medical image processing \cite{Daniel}.

In addition, new applications include satellite image enhancement \cite{Bhand}, noise suppression and enhancement of speech signal \cite{Garg} as well as mobile object tracking \cite{Ljou} and video target tracking \cite{Walia}.

\subsection{Cyber-System and Cryptosystem}

As the importance of cyber-security becomes more crucial, new applications provide good alternatives to conventional approaches. New applications include cryptanalysis \cite{Bhat}, cyber-physical system \cite{Cui}, analysis of LFSR cyptosystem \cite{Din}, vulnerabilities and mitigation optimization in the cloud \cite{Zine}.

\subsection{Sensor Networks and Smart Homes}

As the demand of smart sensors and smart homes increases in modern societies, this becomes an emerging area of application of nature-inspired algorithms. Cuckoo search has been applied to
    node localization in wireless sensor network \cite{Cheng},
    cover optimization of smart home \cite{Sun}, source localization for wireless network \cite{Aziz}, and efficient detection of faulty nodes in distributed systems \cite{Teske}.

There are many other applications such as telecommunications \cite{YangTele}, and readers can refer to Shehab et al. \cite{Shehab} and Yang and Deb \cite{YangDebRev} for more details.

\section{Opportunities}
From the above brief review, we can see that some areas have many applications, while other areas have only a few case studies. Therefore, there are huge opportunities for further research in the following areas.

\begin{itemize}

\item Big Data: Though CS has been applied to data mining, the applications are mainly feature selection and classifications. Further research should focus on the extension of this technique in combination with other techniques such as traditional data mining techniques to do data mining for large datasets more effectively.

\item Biomedical applications: The applications in biomedical applications are really weak, which might be due to the lack of access of real data and high computational costs. Further research is encouraged to do applications in biomedical context, pharmaceutical designs, and large-scale protein structure predictions.

\item NP-hard problems: Though CS has been applied to solve discrete and combinatorial problems such as scheduling and travelling salesman problems, the application to NP-hard problems still have a lot of room for improvement. Travelling salesman problems are still very time-consuming to solve, and the predictions of protein structures are among the most challenging problems.

\item Deep learning: The application of CS in the area of  machine learning and computational intelligence is still weak, which provides a lot of opportunities. For example, the tuning of key parameters in deep nets can be explored by CS and its hybrid variants. Combination of techniques with neural networks and support vector machines can be useful in this area.

\item Large-scale problems: Many case studies in the literature, though with very good results and higher efficiency, are still concerned with small or moderate scale problems where the number of parameters is usually up to a few dozen. Many real-world applications can have thousands of parameters or more and such large-scale problems can pose further challenging for any existing techniques. As the speed of computers continue to increase, this is an area that can be explored in greater detail in the near future.

\end{itemize}

There are other opportunities for further research, including swarming robots, computational intelligence, business applications, transport systems, telecommunications, smart cities, energy-efficient buildings and others. Due to the page limit, we will not discuss them further. It is hoped that this summary can inspire more research in the near future.


\end{document}